\documentclass[final,twocolumn,10pt]{IEEEtran}
\usepackage{graphicx,times,amsmath,amssymb,cite,subfigure,algpseudocode}
\usepackage[ruled]{algorithm}
\IEEEoverridecommandlockouts \allowdisplaybreaks
\DeclareMathOperator*{\argmin}{\arg\!\min}

\floatname{algorithm}{Procedure}

\begin{document}

\title{Online and Offline Domain Adaptation\\ for Reducing BCI Calibration Effort}

\author{\IEEEauthorblockN{Dongrui Wu, \textit{Senior Member, IEEE}}\\
\IEEEauthorblockA{DataNova, NY USA}\\
E-mail: drwu09@gmail.com}

\maketitle

\begin{abstract}
Many real-world brain-computer interface (BCI) applications rely on single-trial classification of event-related potentials (ERPs) in EEG signals. However, because different subjects have different neural responses to even the same stimulus, it is very difficult to build a generic ERP classifier whose parameters fit all subjects. The classifier needs to be calibrated for each individual subject, using some labeled subject-specific data. This paper proposes both online and offline weighted adaptation regularization (wAR) algorithms to reduce this calibration effort, i.e., to minimize the amount of labeled subject-specific EEG data required in BCI calibration, and hence to increase the utility of the BCI system. We demonstrate using a visually-evoked potential oddball task and three different EEG headsets that both online and offline wAR algorithms significantly outperform several other algorithms. Moreover, through source domain selection, we can reduce their computational cost by about $50\%$, making them more suitable for real-time applications.
\end{abstract}


\section{Introduction}

Many real-world brain-computer interface (BCI) applications rely on single-trial classification of event-related potentials (ERPs) in EEG signals \cite{Sajda2010,Bigdely-Shamlo2008}. For example, in a rapid serial visual presentation (RSVP) based BCI system, a sequence of images are shown to the subject rapidly (e.g. 2-10 Hz) \cite{Chun1995,Potter1976}, and the subject needs to detect some target images in them. The target images are much less frequent than the non-target ones, so that they can elicit P300 ERPs in the oddball paradigm. The P300 ERPs can be detected by a BCI system \cite{Pohlmeyer2011}, and the corresponding images are then triaged for further inspection. Research \cite{Sajda2010,Parra2008,Zander2011} has shown that these BCI systems enable the subject to detect targets in large aerial photographs faster and more accurately than traditional standard searches.

Unfortunately, because different subjects have different neural responses to even the same visual stimulus \cite{Kuba2012,Bottger2002,Bulayeva1993}, it is very difficult, if not impossible, to build a generic ERP classifier whose parameters fit all subjects. So, we need to calibrate the classifier for each individual subject, using some labeled subject-specific data. Reducing this calibration effort, i.e., minimizing the amount of labeled subject-specific data required in the calibration, would greatly increase the utility of the BCI system. This is the research problem tackled in this paper.

More specifically, we distinguish between two types of calibration in BCI:
\begin{enumerate}
\item \emph{Offline calibration}, in which a pool of unlabeled EEG epochs have been obtained \emph{a priori}, and a subject is queried to label some of these epochs, which are then used to train a classifier to label the remaining epochs in that pool. A potential application of offline calibration is personalized automatic game highlight detection, e.g., a subject's EEG signals are recorded continuously while watching a football game; after the game, the subject manually labels a few highlights, which are then used to train an ERP classifier to detect more highlights.
\item \emph{Online calibration}, in which some labeled EEG epochs are obtained on-the-fly, and then a classifier is trained from them to classify future  EEG epochs. A potential application of online calibration is the afore-mentioned RSVP image tagging problem: at the beginning of the task, the subject is asked to explicitly indicate it (e.g., press a button) every time he/she detects a target image, and that information is used to train a P300 ERP classifier. After a certain number of calibration epochs, the performance of the classifier can become reliable enough so that it can label further images using EEG epochs only.
\end{enumerate}
One major difference between offline calibration and online calibration is that, in offline calibration, the unlabeled EEG epochs can be used to help design the ERP classifier, whereas in online calibration there are no unlabeled EEG epochs. Additionally, in offline calibration we can query any epoch in the pool for the label (an optimal query strategy can hence be designed by using machine learning methods such as active learning \cite{drwuSMC2014,Settles2009}), but in online calibration usually the sequence of the epochs is pre-determined and the subject has no control over which epoch he/she will see next.

Many signal processing and machine learning approaches have been proposed to reduce the BCI calibration effort \cite{Makeig2012,Lotte2015}. They may be grouped into five categories \cite{Lotte2015}:
\begin{enumerate}
\item \emph{Regularization}, which is a very effective machine learning approach for constructing robust models \cite{Scholkopf2001}, especially when the training data size is small. A popular regularization approach in BCI calibration is shrinkage \cite{Lotte2010}, which gives a regularized estimate of the covariance matrices.
\item \emph{Transfer/multi-task learning}, which uses relevant data from other subjects to help the current subject \cite{Jayaram2016,Wang2015}. The transfer learning (TL) \cite{Pan2010} based approaches are particularly popular \cite{Alamgir2010,Kindermans2012,Samek2013,drwuSMC2014,drwuSMC2015,drwuACII2015,Jayaram2016,Wang2015,drwuPLOS2013}, because in many BCI applications we can easily find legacy data from the same subject in the same task or similar tasks, or legacy data from different subjects in the same task or similar tasks. These data, which will be called auxiliary data in this paper, can be used to improve the learning performance of a new subject, or for a new task.
\item \emph{Adaptive learning}, which refines the machine learning model as new (labeled or unlabeled) subject-specific data are available \cite{Li2008,drwuSMC2015,drwuACII2015}. The main approach in this category is semi-supervised learning \cite{Chapelle2006}, which is often used for offline BCI calibration where unlabeled data are available. Semi-supervised learning first constructs an initial model from the labeled training data and then applies it to the unlabeled test data. The newly labeled test data are then integrated with the groundtruth training data to retrain the model, and hence to improve it iteratively.
\item \emph{Active learning}, which optimally selects the most informative unlabeled samples to label \cite{drwuSMC2014,drwuRSVP2016,drwuEBMAL2016,drwuSMC2015AL}. There are many criteria to determine which unlabeled samples are the most informative \cite{Settles2009}. The most popular, and probably also the simplest, approach for classification is to select the samples that are closest to the current decision boundary, because the classifier is most uncertain about them. Active learning has been mainly used for offline BCI calibration, where unlabeled samples are available. However, a closely related technique, active class selection \cite{Lomasky2007}, can be used for online BCI calibration \cite{drwuPLOS2013}. Its idea is to optimize the classes from which the new training samples are generated.
\item \emph{\emph{A priori} physiological information}, which can be used to construct the most useful EEG features. For example, prior information on which EEG channels are the most likely to be useful was used in \cite{Lotte2011} as a regularizer to optimize spatial filters, and beamforming has been used in \cite{Grosse2009} to find relevant features from prior regions of interest to reduce the calibration data requirement.
\end{enumerate}

Interestingly, these five categories of approaches are not mutually exclusive: in fact they can be freely combined to further reduce the amount of subject-specific calibration data. An optimal spatial filters was designed in \cite{Lotte2011} for efficient subject-to-subject transfer by combining regularization and \emph{a prior} information on which channels are the most likely to be useful. A collaborative filtering approach was developed in \cite{drwuPLOS2013}, which combined TL and active class selection to minimize the online calibration effort. An active TL approach was proposed in \cite{drwuSMC2014} for offline calibration, which combined TL and active learning to minimize the offline calibration effort. An active weighted adaptation regularization approach, which combines active learning, TL, regularization and semi-supervised learning, was proposed in \cite{drwuTNSRE2016} to facilitate the switching between different EEG headsets. All these approaches are for BCI classification problems. However, recently researchers have also started to apply these techniques for regression problems in BCI calibration. For example, a domain adaptation with model fusion approach, which combines regularization and TL, was developed in \cite{drwuaBCI2015} to estimate driver's drowsiness online continuously.

This paper presents a comprehensive overview and comparison of the offline and online weighted adaptation regularization with source domain selection (wARSDS) algorithms, which we proposed recently \cite{drwuSMC2015,drwuACII2015}. The offline wARSDS algorithm, which combined TL, regularization, and semi-supervised learning, was first developed in  \cite{drwuSMC2015} for offline single-trial classification of ERPs in a visually-evoked potential (VEP) oddball task. It was later extended to online calibration in a RSVP task \cite{drwuACII2015}, which still includes TL and regularization but not semi-supervised learning because unlabeled samples are not available in online calibration. In this paper we use a VEP oddball task and three different EEG headsets to show that they have consistently good performance across subjects and headsets. We also compare the performances of the offline and online wARSDS algorithms in identical experimental settings to investigate the effect of semi-supervised learning, and show that it can indeed help improve the calibration performance.

The remainder of the paper is organized as follows: Section~\ref{sect:wARSDS} introduces the details of the offline wARSDS algorithm. Section~\ref{sect:OwARSDS} introduces the online wARSDS (OwARSDS) algorithm. Section~\ref{sect:experiments} describes the experiment setup that is used to evaluate the performances of different algorithms. Section~\ref{sect:offline} presents performance comparison of different offline calibration algorithms. Section~\ref{sect:online} presents performance comparison of different online calibration algorithms. Section~\ref{sect:onoff} compares the performances of offline and online algorithms. Finally, Section~\ref{sect:conclusions} draws conclusions.

\section{Weighted Adaptation Regularization with Source Domain Selection (wARSDS)} \label{sect:wARSDS}

This section describes the offline wARSDS algorithm \cite{drwuSMC2015,drwuTNSRE2016}, which originates from the adaptation regularization -- regularized least squares (ARRLS) algorithm in \cite{Long2014}. We made several major enhancements to ARRLS to better handle class-imbalance and multiple source domains, and also to make use of labeled samples in the target domain. wARSDS first uses source domain selection (SDS) to select the closest source domains for a given target domain, then uses weighted adaptation regularization (wAR) for each selected source domain to build individual classifiers, and finally performs model fusion. For simplicity, we only consider 2-class classification.

\subsection{wAR: Problem Definition}

A \emph{domain} \cite{Pan2010,Long2014} $\mathcal{D}$ in TL consists of a multi-dimensional feature space $\mathcal{X}$ and a marginal probability distribution $P(\mathbf{x})$, i.e., $\mathcal{D}=\{\mathcal{X},P(\mathbf{x})\}$, where $\mathbf{x}\in \mathcal{X}$. Two domains $\mathcal{D}_s$ and $\mathcal{D}_t$ are different if $\mathcal{X}_s\neq \mathcal{X}_t$, and/or $P_s(\mathbf{x})\neq P_t(\mathbf{x})$.

A \emph{task} \cite{Pan2010,Long2014} $\mathcal{T}$ in TL consists of a label space $\mathcal{Y}$ and a conditional probability distribution $Q(y|\mathbf{x})$. Two tasks $\mathcal{T}_s$ and $\mathcal{T}_t$ are different if $\mathcal{Y}_s\neq \mathcal{Y}_t$, or $Q_s(y|\mathbf{x})\neq Q_t(y|\mathbf{x})$.

Given a source domain $\mathcal{D}_s$ with $n$ labeled samples, $\{(\mathbf{x}_1,y_1),..., (\mathbf{x}_n,y_n)\}$, and a target domain $\mathcal{D}_t$ with $m_l$ labeled samples $\{(\mathbf{x}_{n+1},y_{n+1}),...,(\mathbf{x}_{n+m_l},y_{n+m_l})\}$ and $m_u$ unlabeled samples $\{\mathbf{x}_{n+m_l+1}, ..., \mathbf{x}_{n+m_l+m_u}\}$, \emph{domain adaptation} (DA) TL learns a target prediction function $f: \mathbf{x}_t \mapsto y_t$ with low expected error on $\mathcal{D}_t$, under the assumptions $\mathcal{X}_s=\mathcal{X}_t$, $\mathcal{Y}_s=\mathcal{Y}_t$, $P_s(\mathbf{x})\neq P_t(\mathbf{x})$, and $Q_s(y|\mathbf{x})\neq Q_t(y|\mathbf{x})$.

For example, in single-trial classification of VEPs, EEG epochs from the new subject are in the target domain, while EEG epochs from an existing subject (usually different from the new subject) are in the source domain. When there are multiple source domains, we perform DA for each of them separately and then aggregate the classifiers. A sample consists of the feature vector for an EEG epoch from a subject in either domain, collected as a response to a specific visual stimulus. Though usually the source and target domains employ the same feature extraction method, generally their marginal and conditional probability distributions are different, i.e., $P_s(\mathbf{x})\neq P_t(\mathbf{x})$ and $Q_s(y|\mathbf{x})\neq Q_t(y|\mathbf{x})$, because the two subjects usually have different neural responses to the same visual stimulus \cite{Kuba2012,Bottger2002,Bulayeva1993}. As a result, the auxiliary data from a source domain cannot represent the primary data in the target domain accurately, and must be integrated with some labeled target domain data to induce an accurate target domain classifier.

\subsection{wAR: The Learning Framework}

Because
\begin{align}
f(\mathbf{x})=Q(y|\mathbf{x})=\frac{P(\mathbf{x},y)}{P(\mathbf{x})}=\frac{Q(\mathbf{x}|y)P(y)}{P(\mathbf{x})},
\end{align}
to use the source domain data in the target domain, we need to make sure $P_s(\mathbf{x}_s)$ is close to $P_t(\mathbf{x}_t)$, $Q_s(\mathbf{x}_s|y_s)$ is close to $Q_t(\mathbf{x}_t|y_t)$, and $P_s(y)$ is also close to $P_t(y)$. However, in this paper we focus only on the first two requirements by assuming all subjects conduct similar VEP tasks [so $P_s(y)$ and $P_t(y)$ are intrinsically close]. Our future research will consider the more general case that $P_s(y)$ and $P_t(y)$ are different.

Let the classifier be $f(\mathbf{x})=\mathbf{w}^T\phi(\mathbf{x})$, where $\mathbf{w}$ is the classifier parameters, and $\phi:\mathcal{X}\mapsto \mathcal{H}$ is the feature mapping function that projects the original feature vector to a Hilbert space $\mathcal{H}$. As in\cite{drwuSMC2015,drwuTNSRE2016}, the learning framework of wAR is formulated as:
\begin{align}
f=&\argmin\limits_{f\in\mathcal{H}_K}\sum_{i=1}^{n}w_{s,i}\ell(f(\mathbf{x}_i),y_i)+w_t\sum_{i=n+1}^{n+m_l}w_{t,i}\ell(f(\mathbf{x}_i),y_i)\nonumber \\
&+\sigma\|f\|_K^2+\lambda [D_{f,K}(P_s,P_t)+ D_{f,K}(Q_s,Q_t)] \label{eq:f}
\end{align}
where $\ell$ is the loss function, $K\in R^{(n+m_l+m_u)\times(n+m_l+m_u)}$ is the kernel matrix with $K(\mathbf{x}_i,\mathbf{x}_j)=\langle\phi(\mathbf{x}_i),\phi(\mathbf{x}_j)\rangle$, and $\sigma$ and $\lambda$ are non-negative regularization parameters. $w_t$ is the overall weight for target domain samples, which should be larger than 1 so that more emphasis is given to target domain samples than source domain samples\footnote{Generally the number of labeled samples in the source domain is much larger than that in the target domain in both offline and online calibration scenarios, i.e., $n \gg m_l$. However, eventually the learned classifier will be applied to the target domain. So, the target domain should be emphasized. We choose $w_t>1$ so that the $m_l$ labeled target domain samples are less overwhelmed by the $n$ source domain samples.}. $w_{s,i}$ and $w_{t,i}$ are the weight for the $i^\mathrm{th}$ sample in the source domain and target domain, respectively, i.e.,
\begin{align}
w_{s,i}&=\left\{\begin{array}{ll}
                  1, & \mathbf{x}_i\in \mathcal{D}_{s,1} \\
                  n_1/(n-n_1), & \mathbf{x}_i\in \mathcal{D}_{s,2}
                \end{array}\right. \label{eq:ws}\\
w_{t,i}&=\left\{\begin{array}{ll}
                  1, & \mathbf{x}_i\in \mathcal{D}_{t,1} \\
                  m_1/(m_l-m_1), & \mathbf{x}_i\in \mathcal{D}_{t,2}
                \end{array}\right.  \label{eq:wt}
\end{align}
in which $\mathcal{D}_{s,c}=\{\mathbf{x}_i|\mathbf{x}_i\in \mathcal{D}_s\wedge y_i=c,\ i=1,...,n\}$ is the set of samples in Class $c$ of the source domain, $\mathcal{D}_{t,c}=\{\mathbf{x}_j|\mathbf{x}_j\in \mathcal{D}_t\wedge y_j=c,\ j=n+1,...,n+m_l\}$ is the set of samples in Class $c$ of the target domain, $n_c$ is the number of elements in $\mathcal{D}_{s,c}$, and $m_c$ is the number of elements in $\mathcal{D}_{t,c}$. The goal of $w_{s,i}$ ($w_{t,i}$) is to balance the number of samples from difference classes in the source (target) domain. This is very important, because class imbalance is intrinsic to many applications \cite{Japkowicz2002}, particularly BCI applications. In many cases the minority class is the one of particular interest (e.g., the VEP experiment presented in this paper), but it can be easily overwhelmed by the majority class if not properly weighted. Of course, there are many other approaches for handling class imbalance \cite{Longadge2013,Provost2000,Japkowicz2002}. We used the weighting approach for its simplicity.

Briefly speaking, the 1st term in (\ref{eq:f}) minimizes the loss on fitting the labeled samples in the source domain, the 2nd term minimizes the loss on fitting the labeled samples in the target domain, the 3rd term minimizes the structural risk of the classifier, and the 4th term minimizes the distance between the marginal probability distributions $P_s(\mathbf{x}_s)$ and $P_t(\mathbf{x}_t)$, and also the distance between the conditional probability distributions $Q_s(\mathbf{x}_s|y_s)$ and $Q_t(\mathbf{x}_t|y_t)$.

According to the Representer Theorem \cite{Belkin2006,Long2014}, the solution of (\ref{eq:f}) can be expressed as:
\begin{align}
f(\mathbf{x})=\sum_{i=1}^{n+m_l+m_u}\alpha_iK(\mathbf{x}_i,\mathbf{x})=\boldsymbol{\alpha}^TK(X,\mathbf{x}) \label{eq:f2}
\end{align}
where
\begin{align}
X=[\mathbf{x}_1, ...,\mathbf{x}_{n+m_l+m_u}]^T \label{eq:X}
\end{align}
and $\boldsymbol{\alpha}=[\alpha_1,...,\alpha_{n+m_l+m_u}]^T$ are coefficients to be computed.


\subsection{wAR: Loss Functions Minimization}

Let
\begin{align}
\mathbf{y}=[y_1,...,y_{n+m_l+m_u}]^T \label{eq:y}
\end{align}
where $\{y_1,...,y_n\}$ are known labels in the source domain, $\{y_{n+1},...,y_{n+m_l}\}$ are known labels in the target domain, and $\{y_{n+m_l+1},...,y_{n+m_l+m_u}\}$ are pseudo labels for the unlabeled target domain samples, i.e., labels estimated using available sample information in both source and target domains.

Define $E\in R^{(n+m_l+m_u)\times(n+m_l+m_u)}$ as a diagonal matrix with
\begin{align}
E_{ii}=\left\{\begin{array}{ll}
                w_{s,i}, & 1\le i\le n \\
                w_tw_{t,i}, &  n+1 \le i \le n+m_l\\
                0, & \text{otherwise}
              \end{array}\right. \label{eq:E}
\end{align}

We use the squared loss in this paper:
\begin{align}
\ell(f(\mathbf{x}_i),y_i)=(y_i-f(\mathbf{x}_i))^2 \label{eq:l2}
\end{align}

Substituting (\ref{eq:f2}) and (\ref{eq:l2}) into the first two terms in (\ref{eq:f}), we have
\begin{align}
&\sum_{i=1}^{n}w_{s,i}\ell(f(\mathbf{x}_i),y_i)+w_t\sum_{i=n+1}^{n+m_l}w_{t,i}\ell(f(\mathbf{x}_i),y_i)\nonumber\\
=&\sum_{i=1}^{n}w_{s,i}(y_i-f(\mathbf{x}_i))^2+w_t\sum_{i=n+1}^{n+m_l}w_{t,i}(y_i-f(\mathbf{x}_i))^2 \nonumber \\
=&\sum_{i=1}^{n+m_l+m_u}E_{ii}(y_i-f(\mathbf{x}_i))^2 \nonumber \\
=&(\mathbf{y}^T-\boldsymbol{\alpha}^TK)E(\mathbf{y}-K\boldsymbol{\alpha}) \label{eq:l3}
\end{align}

\subsection{wAR: Structural Risk Minimization}

As in \cite{Long2014}, we define the structural risk as the squared norm of $f$ in $\mathcal{H}_K$, i.e.,
\begin{align}
\|f\|_K^2=\sum_{i=1}^{n+m_l+m_u} \sum_{j=1}^{n+m_l+m_u}\alpha_i\alpha_jK(\mathbf{x}_i,\mathbf{x}_j)=\boldsymbol{\alpha}^T K\boldsymbol{\alpha} \label{eq:fK}
\end{align}

\subsection{wAR: Marginal Probability Distribution Adaptation}

As in \cite{Long2014}, we compute $D_{f,K}(P_s,P_t)$ using the projected maximum mean discrepancy (MMD) between the source and target domains:
\begin{align}
D_{f,K}(P_s,P_t)&=\left[\frac{1}{n}\sum_{i=1}^nf(\mathbf{x}_i)-\frac{1}{m_l+m_u}\sum_{i=n+1}^{n+m_l+m_u}f(\mathbf{x}_i)\right]^2 \nonumber\\
&=\boldsymbol{\alpha}^TKM_0K\boldsymbol{\alpha} \label{eq:DfKP}
\end{align}
where $M_0\in R^{(n+m_l+m_u)\times (n+m_l+m_u)}$ is the MMD matrix:
\begin{align}
(M_0)_{ij}=\left\{\begin{array}{ll}
                             \frac{1}{n^2},& 1\le i \le n, 1\le j \le n  \\
                             \frac{1}{(m_l+m_u)^2}, & n+1\le i \le n+m_l+m_u,\\
                             & n+1\le j \le n+m_l+m_u \\
                             \frac{-1}{n(m_l+m_u)}, & \text{otherwise}
                           \end{array}\right. \label{eq:M0}
\end{align}

\subsection{wAR: Conditional Probability Distribution Adaptation}

As in \cite{Long2014}, we first compute pseudo labels for the unlabeled target domain samples and construct the label vector $\mathbf{y}$ in (\ref{eq:y}). These pseudo labels can be computed using the classifier built in the previous iteration if wAR is used iteratively, or estimated using another classifier, e.g., a support vector machine (SVM) \cite{Vapnik1998}. We then compute the projected MMD with respect to each class.

Let $\mathcal{D}_{s,c}=\{\mathbf{x}_i|\mathbf{x}_i\in \mathcal{D}_s\wedge y_i=c,\ i=1,...,n\}$ be the set of samples in Class $c$ of the source domain, $\mathcal{D}_{t,c}=\{\mathbf{x}_j|\mathbf{x}_j\in \mathcal{D}_t\wedge y_j=c,\ j=n+1,...,n+m_l+m_u\}$ be the set of samples in Class $c$ of the target domain, $n_c$ be the number of elements in $\mathcal{D}_{s,c}$, and $m_c$ be the number of elements in $\mathcal{D}_{t,c}$. Then, the distance between the conditional probability distributions in the two domains is computed as:
\begin{align}
D_{f,K}(Q_s,Q_t)
=\sum_{c=1}^2\left[\frac{1}{n_c}\sum\limits_{\mathbf{x}_i\in \mathcal{D}_{s,c}} f(\mathbf{x}_i)-\frac{1}{m_c}\sum\limits_{\mathbf{x}_j\in\mathcal{D}_{t,c}}f(\mathbf{x}_j)\right]^2 \label{eq:DfKQ}
\end{align}

Substituting (\ref{eq:f2}) into (\ref{eq:DfKQ}), it follows that
\begin{align}
&D_{f,K}(Q_s,Q_t)\nonumber \\
=&\sum_{c=1}^2\left[\frac{1}{n_c}\sum\limits_{\mathbf{x}_i\in \mathcal{D}_{s,c}} \boldsymbol{\alpha}^TK(X,\mathbf{x}) -\frac{1}{m_c}\sum\limits_{\mathbf{x}_j\in\mathcal{D}_{t,c}}\boldsymbol{\alpha}^TK(X,\mathbf{x})\right]^2 \nonumber\\
=&\sum_{c=1}^2\boldsymbol{\alpha}^TKM_cK\boldsymbol{\alpha} =\boldsymbol{\alpha}^TKMK\boldsymbol{\alpha} \label{eq:DfKQ2}
\end{align}
where
\begin{align}
M=M_1+M_2 \label{eq:M}
\end{align}
in which $M_1$ and $M_2$ are MMD matrices computed as:
\begin{align}
(M_c)_{ij}=\left\{\begin{array}{ll}
                    1/n_c^2, & \mathbf{x}_i, \mathbf{x}_j\in \mathcal{D}_{s,c} \\
                    1/m_c^2, & \mathbf{x}_i, \mathbf{x}_j\in \mathcal{D}_{t,c} \\
                    -1/(n_cm_c), & \mathbf{x}_i\in\mathcal{D}_{s,c}, \mathbf{x}_j\in \mathcal{D}_{t,c}, \\
                    &\text{or } \mathbf{x}_j\in\mathcal{D}_{s,c}, \mathbf{x}_i\in \mathcal{D}_{t,c} \\
                    0, & \text{otherwise}
                  \end{array}\right.
\end{align}

\subsection{wAR: The Closed-Form Solution}

Substituting (\ref{eq:l3}), (\ref{eq:fK}), (\ref{eq:DfKP}) and (\ref{eq:DfKQ2}) into (\ref{eq:f}), we have
\begin{align}
f=\argmin\limits_{f\in\mathcal{H}_K} &(\mathbf{y}^T-\boldsymbol{\alpha}^TK)E(\mathbf{y}-K\boldsymbol{\alpha}) \nonumber \\
& + \sigma \boldsymbol{\alpha}^T K\boldsymbol{\alpha} +  \lambda\boldsymbol{\alpha}^TK(M_0+M)K\boldsymbol{\alpha} \label{eq:f3}
\end{align}
Setting the derivative of the objective function above to 0 leads to the following closed-form solution for $\boldsymbol{\alpha}$:
\begin{align}
\boldsymbol{\alpha}=[(E+\lambda M_0+ \lambda M)K+\sigma I]^{-1}E\mathbf{y} \label{eq:alpha}
\end{align}

\subsection{Source Domain Selection (SDS)}

When there are multiple source domains, it is very time-consuming to perform wAR for each source domain and then aggregate the results. Additionally, aggregating results from source domains that are outliers or very different from the target domain may also hurt the classification performance. So, we introduce a source domain selection approach \cite{drwuSMC2015}, which selects the closest source domains to reduce the computational cost, and also to (potentially) improve the classification performance.

Assume there are $Z$ different source domains. For the $z^\mathrm{th}$ source domain, we first compute $\mathbf{m}_{z,c}$ ($c=1,2$), the mean feature vector of each class. Then, we also compute $\mathbf{m}_{t,c}$, the mean feature vector of each target domain class, by making use of the $m_l$ known labels and the $m_u$ pseudo-labels. The distance between the two domains is then computed as:
\begin{align}
d(z,t)=\sum_{c=1}^2 \|\mathbf{m}_{z,c}-\mathbf{m}_{t,c}\| \label{eq:dST}
\end{align}
We next cluster these $Z$ distances, $\{d(z,t)\}_{z=1,...,Z}$, by $k$-means clustering, and finally choose the cluster that has the smallest centroid, i.e., the source domains that are closest to the target domain. In this way, on average we only need to perform wAR for $Z/k$ ($k$ is the number of clusters in $k$-means clustering) source domains, corresponding to a 50\% computational cost saving if $k=2$. A larger $k$ will result in a larger saving; however, when $k$ is too large, there may not be enough source domains selected for wAR, and hence the classification performance may be unstable. So, there is a trade-off between computational cost saving and classification performance. $k=2$ was used in this paper, and it demonstrated satisfactory performance.

\subsection{The Complete wARSDS Algorithm}

The pseudo code for the complete wARSDS algorithm is shown in Algorithm~1. It first uses SDS to select the closest source domains, then performs wAR for each of them separately to build individual classifiers, and finally aggregates them using a weighted average, where the weights are the corresponding training accuracies.

\subsection{Discussions}

As mentioned at the beginning of this section, the formulation and derivation of wAR closely resemble the ARRLS algorithm in \cite{Long2014}; however, there are several major differences:
\begin{enumerate}
\item wAR assumes a subject or an oracle is available to label a small number of samples in the target domain, whereas ARRLS assumes all target domain samples are unlabeled. As a result, wAR can be iterative, and the classifier can be updated every time new labeled target domain samples are available.
\item wAR explicitly considers the class imbalance problem in both source and target domains by introducing the class-dependent weights on samples. As it will be shown in Section~\ref{sect:experiments}, this makes a huge difference in the balanced classification accuracy for the class imbalance problem, which is intrinsic in ERP-based BCI systems.
\item ARRLS also includes manifold regularization \cite{Belkin2006}. We investigated it but was not able to observe improved performance in our application, so it is not included in this paper.
\end{enumerate}
Finally, when combined with SDS, wARSDS can effectively handle multiple source domains, whereas ARRLS only considers one source domain.

\begin{figure*}   \centering
  \begin{minipage}[t]{0.48\textwidth}\small
\alglanguage{pseudocode}
\renewcommand\figurename{Algorithm}
\setcounter{figure}{0}
  \caption{The offline wARSDS algorithm \cite{drwuSMC2015,drwuTNSRE2016}.}  \hrulefill
\begin{algorithmic}
\Require {$Z$ source domains, where the $z^\mathrm{th}$ ($z=1,...,Z$) domain has $n_z$ labeled samples $\{\mathbf{x}_i^z,y_i^z\}_{i=1,...,n_z}$; \\
$m_l$ labeled target domain samples, $\{\mathbf{x}_j^t,y_j^t\}_{j=1,...,m_l}$; \\
$m_u$ unlabeled target domain samples,
$\{\mathbf{x}_j^t\}_{j=m_l+1,...,m_l+m_u}$; \\
Parameters $w_t$, $\sigma$ and $\lambda$ in (\ref{eq:f}); \\
Parameter $k$ in $k$-means clustering of SDS.}
\Ensure The wARSDS classifier $f(\mathbf{x})$.
\Statex \Comment{\texttt{SDS starts}}
\If {$m_l==0$}
\State Retain all $Z$ source domains;
\State Compute pseudo-labels $\{y_j\}_{j=n_z+m_l+1,...,n_z+m_l+m_u}$ using another classifier, e.g., an SVM;
\State Go to wAR.
\Else
\State Compute pseudo-labels $\{y_j\}_{j=n_z+m_l+1,...,n_z+m_l+m_u}$ using the wARSDS classifier built from the previous iteration;
\For{$z=1,2,...,Z$}
\State Compute $d(z,t)$, the distance between the target domain and the $z^\mathrm{th}$ source domain, by (\ref{eq:dST}).
\EndFor
\State Cluster $\{d(z,t)\}_{z=1,...,Z}$ by $k$-means clustering;
\State Retain the $Z'$ source domains that belong to the cluster with the smallest centroid.
\EndIf
\Statex \Comment{\texttt{SDS ends; wAR starts}}
\State Choose a kernel function $K(\mathbf{x}_i,\mathbf{x}_j)$;
\For{$z=1,2,...,Z'$}
\State Construct the feature matrix $X$ in (\ref{eq:X});
\State Compute the kernel matrix $K_z$ from $X$;
\State Construct $\mathbf{y}$ in (\ref{eq:y}), $E$ in (\ref{eq:E}), $M_0$ in (\ref{eq:M0}), and $M$ in (\ref{eq:M});
\State Compute $\boldsymbol{\alpha}$ by (\ref{eq:alpha}) and record it as $\boldsymbol{\alpha}_z$;
\State Use $\boldsymbol{\alpha}$ to classify the $n_z+m_l$ labeled samples and record the accuracy, $w_z$;
\EndFor
\Statex \Comment{\texttt{wAR ends; Aggregation starts}}\\
\Return $f(\mathbf{x})=\sum_{z=1}^{Z'}w_z\boldsymbol{\alpha}_zK_z(X,\mathbf{x})$.\\
\hrulefill
\end{algorithmic}
  \end{minipage}\hfill
  \begin{minipage}[t]{0.48\textwidth}\small
  \alglanguage{pseudocode}
  \renewcommand\figurename{Algorithm}
\setcounter{figure}{1}
\caption{The online OwARSDS algorithm \cite{drwuACII2015}.} \label{alg:OwARSDS}  \hrulefill
\begin{algorithmic}
\Require {$Z$ source domains, where the $z^\mathrm{th}$ ($z=1,...,Z$) domain has $n_z$ labeled samples $\{\mathbf{x}_i^z,y_i^z\}_{i=1,...,n_z}$;\\
$m_l$ labeled target domain samples, $\{\mathbf{x}_j^t,y_j^t\}_{j=1,...,m_l}$;\\
\Statex
Parameters $w_t$, $\sigma$ and $\lambda$ in (\ref{eq:f});\\
Parameter $k$ in $k$-means clustering of SDS.}
\Ensure {The OwARSDS classifier $f^o(\mathbf{x})$.}
\Statex \Comment{\texttt{SDS starts}}
\If {$m_l==0$}
\State Retain all $Z$ source domains;
\Statex
\Statex
\State Go to OwAR.
\Else
\Statex
\Statex
\For{$z=1,2,...,Z$}
\State Compute $d(z,t)$, the distance between the target domain and the $z^\mathrm{th}$ source domain, by (\ref{eq:dST}).
\EndFor
\State Cluster $\{d(z,t)\}_{z=1,...,Z}$ by $k$-means clustering;
\State Retain the $Z'$ source domains that belong to the cluster with the smallest centroid.
\EndIf
\Statex \Comment{\texttt{SDS ends; OwAR starts}}\\
Choose a kernel function $K^o(\mathbf{x}_i,\mathbf{x}_j)$ ;
\For{$z=1,2,...,Z'$}
\State Construct the feature matrix $X^o$ in (\ref{eq:Xo});
\State Compute the kernel matrix $K_z^o$ from $X^o$;
\State Construct $\mathbf{y}^o$ in (\ref{eq:yonline}), $E^o$ in (\ref{eq:Eonline}), $M_0^o$ in (\ref{eq:M0online}), $M^o$ in (\ref{eq:DfKQ2online});
\State Compute $\boldsymbol{\alpha}^o$ by (\ref{eq:alpha}) and record it as $\boldsymbol{\alpha}_z^o$;
\State Use $\boldsymbol{\alpha}_z^o$ to classify the $n_z+m_l$ labeled samples and record the accuracy, $w_z^o$;
\EndFor
\Statex \Comment{\texttt{OwAR ends; Aggregation starts}}\\
\Return $f^o(\mathbf{x})=\sum_{z=1}^{Z'}w_z^o\boldsymbol{\alpha}_z^oK_z^o(X^o,\mathbf{x})$.\\
 \hrulefill
\end{algorithmic}
  \end{minipage}
\end{figure*}
\setcounter{figure}{0}

\section{Online Weighted Adaptation Regularization with Source Domain Selection (OwARSDS)} \label{sect:OwARSDS}

This section introduces the OwARSDS algorithm \cite{drwuACII2015}, which extends the offline wARSDS algorithm to online BCI calibration. OwARSDS first uses SDS to select the closest source domains, then performs online weighted adaptation regularization (OwAR) for each selected source domain to build individual classifiers, and finally aggregates them.

\subsection{OwAR: The Learning Framework}

Using the notations introduced in the previous section, the learning framework of OwAR can still be formulated as (\ref{eq:f}). However, because in online calibration there are no unlabeled target domain samples, the kernel matrix $K^o$ has dimensionality $(n+m_l)\times(n+m_l)$, instead of $(n+m_l+m_u)\times(n+m_l+m_u)$ in offline calibration. As a result, the solution of (\ref{eq:f}) admits a different expression:
\begin{align}
f^o(\mathbf{x})=\sum_{i=1}^{n+m_l}\alpha_i^oK^o(\mathbf{x}_i,\mathbf{x})=(\boldsymbol{\alpha}^o)^TK^o(X^o,\mathbf{x}) \label{eq:fonline}
\end{align}
where
\begin{align}
X^o=[\mathbf{x}_1, ...,\mathbf{x}_{n+m_l}]^T  \label{eq:Xo}
\end{align}
and $\boldsymbol{\alpha}^o=[\alpha_1,...,\alpha_{n+m_l}]^T$ are coefficients to be computed.

It has been shown \cite{drwuACII2015} that the closed-form solution for $\boldsymbol{\alpha}^o$ is:
\begin{align}
\boldsymbol{\alpha}^o=[(E^o+\lambda M_0^o+ \lambda M^o)K^o+\sigma I]^{-1}E^o\mathbf{y}^o \label{eq:alphaOnline}
\end{align}
Next we briefly introduce how the various terms in (\ref{eq:alphaOnline}) are derived.

\subsection{OwAR: Loss Functions Minimization}

Define
\begin{align}
\mathbf{y}^o=[y_1,...,y_{n+m_l}]^T \label{eq:yonline}
\end{align}
where $\{y_1,...,y_n\}$ are known labels in the source domain, and $\{y_{n+1},...,y_{n+m_l}\}$ are known labels in the target domain. Define also $E^o\in R^{(n+m_l)\times(n+m_l)}$ as a diagonal matrix with
\begin{align}
E_{ii}^o=\left\{\begin{array}{ll}
                w_{s,i}, & 1\le i\le n \\
                w_tw_{t,i}, &  n+1 \le i \le n+m_l
              \end{array}\right. \label{eq:Eonline}
\end{align}
Then, following the derivation in (\ref{eq:l3}), we now have
\begin{align}
&\sum_{i=1}^{n}w_{s,i}\ell(f^o(\mathbf{x}_i),y_i)+w_t\sum_{i=n+1}^{n+m_l}w_{t,i}\ell(f^o(\mathbf{x}_i),y_i)\nonumber \\
=&[(\mathbf{y}^o)^T-(\boldsymbol{\alpha}^o)^TK^o]E^o(\mathbf{y}^o-K^o\boldsymbol{\alpha}^o) \label{eq:l3online}
\end{align}

\subsection{OwAR: Structural Risk Minimization}

Again, we define the structural risk as the squared norm of $f^o$ in $\mathcal{H}_K$, i.e.,
\begin{align}
\|f^o\|_K^2=(\boldsymbol{\alpha}^o)^T K^o\boldsymbol{\alpha}^o \label{eq:fKonline}
\end{align}

\subsection{OwAR: Marginal Probability Distribution Adaptation}

We compute $D_{f^o,K^o}(P_s,P_t)$ using the projected MMD between the source and target domains:
\begin{align}
D_{f^o,K^o}(P_s,P_t)&=\left[\frac{1}{n}\sum_{i=1}^nf^o(\mathbf{x}_i)-\frac{1}{m_l}\sum_{i=n+1}^{n+m_l}f^o(\mathbf{x}_i)\right]^2 \nonumber \\ &=(\boldsymbol{\alpha}^o)^TK^oM_0^oK^o\boldsymbol{\alpha}^o \label{eq:DfKPonline}
\end{align}
where $M_0^o\in R^{(n+m_l)\times (n+m_l)}$ is the MMD matrix:
\begin{align}
(M_0^o)_{ij}=\left\{\begin{array}{ll}
                             \frac{1}{n^2},& 1\le i \le n, 1\le j \le n  \\
                             \frac{1}{m_l^2}, & n+1\le i \le n+m_l,\\
                             & n+1\le j \le n+m_l \\
                             \frac{-1}{nm_l}, & \text{otherwise}
                           \end{array}\right. \label{eq:M0online}
\end{align}

\subsection{OwAR: Conditional Probability Distribution Adaptation}

In offline calibration, to minimize the discrepancy between the conditional probability distributions in the source and target domains, we need to first compute the pseudo-labels for the $m_u$ unlabeled target domain samples. In online calibration, because there are no unlabeled target domain samples, this step is skipped. Following the derivation of (\ref{eq:DfKQ2}), we still have:
\begin{align}
D_{f^o,K^o}(Q_s,Q_t)=(\boldsymbol{\alpha}^o)^TK^oM^oK^o\boldsymbol{\alpha}^o \label{eq:DfKQ2online}
\end{align}
where $M^o$ is still computed by (\ref{eq:M}), but using only the $n$ source domain samples and $m_l$ target domain samples.

\subsection{Source Domain Selection (SDS)}

The SDS procedure in OwARSDS is almost identical to that in wARSDS. The only difference is that the latter also makes use of the $m_u$ unlabeled target domain samples in computing $\mathbf{m}_{t,c}$ in (\ref{eq:dST}), whereas the former only uses the $m_l$ labeled target domain samples, because there are no unlabeled target domain samples in online calibration.

\subsection{The Complete OwARSDS Algorithm}

The pseudo code for the complete OwARSDS algorithm is described in Algorithm~2. It first uses SDS to select the closest source domains, then performs OwAR for each of them separately to build individual classifiers, and finally aggregates them using a weighted average, where the weights are the corresponding training accuracies. Observe that OwARSDS is very similar to wARSDS, the major difference being that no unlabeled target domain samples are available for use in OwARSDS.


\section{The VEP Oddball Experiment} \label{sect:experiments}

This section describes the setup of the VEP oddball experiment, which is used in the following three sections to evaluate the performances of different algorithms.

\subsection{Experiment Setup}

\begin{figure}[htpb]\centering
\subfigure[]{\label{fig:T}     \includegraphics[width=.20\linewidth]{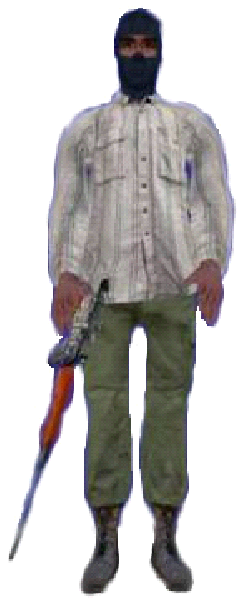}}
\subfigure[]{\label{fig:NT}     \includegraphics[width=.18\linewidth]{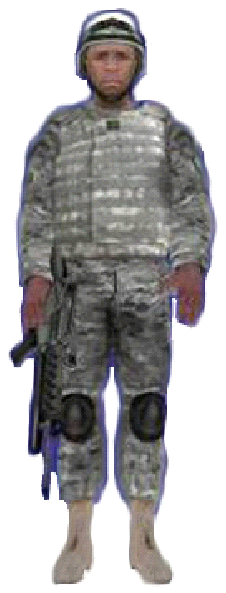}}
\caption{Example images of (a) a target; (b) a non-target.} \label{fig:TNT}
\end{figure}

A two-stimulus VEP oddball task was used \cite{Ries2014}. In this task, participants were seated in a sound-attenuated recording chamber, and image stimuli were presented to them at a rate of 0.5 Hz (one image every two seconds). The images (152$\times$375 pixels), presented for 150 ms at the center of a 24 inch Dell P2410 monitor at a distance of approximately 70 cm, were either an enemy combatant [\emph{target}; an example is shown in Fig.~\ref{fig:T}] or a U.S. Soldier [\emph{non-target}; an example is shown in Fig.~\ref{fig:NT}]. The subjects were instructed to maintain fixation on the center of the screen and identify each image as being target or non-target with a unique button press as quickly and accurately as possible\footnote{In the traditional oddball paradigm, subjects are only asked to respond to the target (oddball) stimuli. In our experiment we asked the subjects to respond to both types of stimuli so that we can remove epochs with incorrect responses from our analysis. Additionally, the experimental data will enable other analyses including the response time to different types of stimuli. Similar experimental settings have been used in \cite{Huettel2004,Herbette2006}.}. A total of 270 images were presented to each subject, among which 34 were targets. The experiments were approved by U.S. Army Research Laboratory (ARL) Institutional Review Board. The voluntary, fully informed consent of the persons used in this research was obtained as required by federal and Army regulations \cite{USArmy,USDoD}. The investigator has adhered to Army policies for the protection of human subjects.

18 subjects participated the experiments, which lasted on average 15 minutes. Signals for each subject were recorded with three different EEG headsets, including a 64-channel 512Hz BioSemi ActiveTwo system, a 9-channel 256Hz Advanced Brain Monitoring (ABM) X10 system, and a 14-channel 128Hz Emotiv EPOC headset. However, due to some exceptions at the experiment, data were correctly recorded for only 16 subjects for ABM, 15 subjects for BioSemi, and 15 subjects for Emotiv. There were 14 subjects whose data were correctly recorded for all three headsets, so we used only these 14 subjects in this study.

\subsection{Preprocessing and Feature Extraction}

The preprocessing and feature extraction method for all three headsets was the same, except that for ABM and Emotiv headsets we used all the channels, but for the BioSemi headset we only used 21 channels (Cz, Fz, P1, P3, P5, P7, P9, PO7, PO3, O1, Oz, POz, Pz, P2, P4, P6, P8, P10, PO8, PO4, O2) mainly in the parietal and occipital areas, as in \cite{drwuSMC2014}.

EEGLAB \cite{Delorme2004} was used for EEG signal preprocessing and feature extraction. For each headset, we first band-passed the EEG signals to [1, 50] Hz, then downsampled them to 64 Hz, performed average reference, and next epoched them to the $[0, 0.7]$ second interval timelocked to stimulus onset. We removed mean baseline from each channel in each epoch and removed epochs with incorrect button press responses\footnote{Button press responses were not recorded for most subjects using the ABM headset, so we used all 270 epochs for them.}. The final numbers of epochs from the 14 subjects are shown in Table~\ref{tab:epoch}. Observe that there is significant class imbalance for every subject; that's why we need to use $w_{s,i}$ and $w_{t,i}$ in (\ref{eq:f}) to balance the two classes in both domains.

\begin{table*}[htpb] \centering \setlength{\tabcolsep}{1mm}
\caption{Number of epoches for each subject after preprocessing. The numbers of target epochs are given in the parentheses.}   \label{tab:epoch}
\begin{tabular}{l|cccccccccccccc}   \hline
   Subject  &  1&2&3&4&5&6&7&8&9&10&11&12&13 &14 \\ \hline
   BioSemi &  241 (26)&260 (24)& 257 (24) & 261 (29)& 259 (29)& 264 (30)& 261 (29) & 252 (22)& 261 (26)& 259 (29)& 267 (32)& 259 (24)&261 (25)& 269 (33)\\
  Emotiv  &263 (28) &  265 (30) &  266 (30)& 255 (23)& 264 (30)& 263 (32)& 266 (30)&252 (22)& 261 (26)& 266 (29)& 266 (32)& 264 (33) & 261 (26)& 267 (31)\\
  ABM & 270 (34) & 270 (34) & 235 (30) & 270 (34) & 270 (34)&270 (34)&270 (34)&270 (33)&270 (34)&239 (30)&270 (34)&270 (34)&251 (31)&270 (34)\\   \hline
\end{tabular}
\end{table*}

Each [0, 0.7] second epoch contains hundreds of raw EEG magnitude samples (e.g., $64\times0.7\times21=924$ for BioSemi). To reduce the dimensionality, we performed a simple principal component analysis (PCA) to take the scores on the first 20 principal components as features. We then normalized each feature dimension separately to $[0, 1]$.

\subsection{Performance Measure}

Let $m_+$ and $m_-$ be the true number of epochs from the target and non-target class, respectively. Let $\hat{m}_+$ and $\hat{m}_-$ be the number of epochs that are correctly classified by an algorithm as target and non-target, respectively. Then, we compute
\begin{align*}
a_+=\frac{\hat{m}_+}{m_+},\qquad a_-=\frac{\hat{m}_-}{m_-}
\end{align*}
where $a_+$ is the classification accuracy on the target class, and $a_-$ is the classification accuracy on the non-target class.

The following balanced classification accuracy (BCA) was then used as the performance measure in this paper:
\begin{align}
BCA=\frac{a_++a_-}{2}. \label{eq:a}
\end{align}

\section{Performance Evaluation of Offline Calibration Algorithms} \label{sect:offline}

This selection presents performance comparison of wARSDS with several other offline calibration algorithms.

\subsection{Calibration Scenario}

Although we knew the labels of all EEG epochs for all 14 subjects in the VEP experiment, we simulated a realistic offline calibration scenario: we had labeled EEG epochs from 13 subjects, and also all epochs from the 14th subject, but initially none of them was labeled. Our goal was to iteratively label epochs from the 14th subject and build a classifier so that his/her remaining unlabeled epochs can be reliably classified.

\begin{figure}[htpb]\centering
\subfigure[]{\label{fig:offlineC}\includegraphics[width=.49\linewidth]{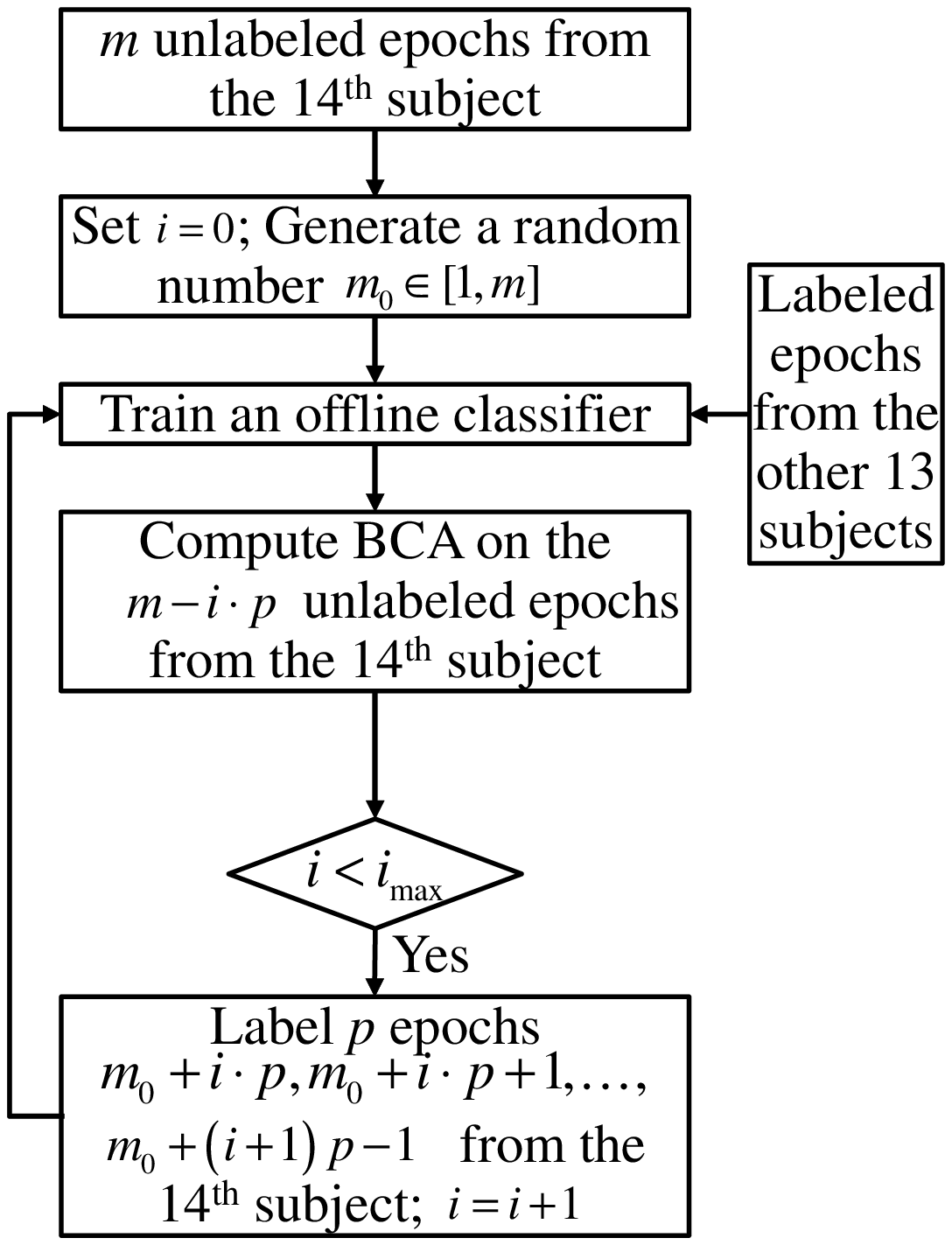}}
\subfigure[]{\label{fig:onlineC}\includegraphics[width=.49\linewidth]{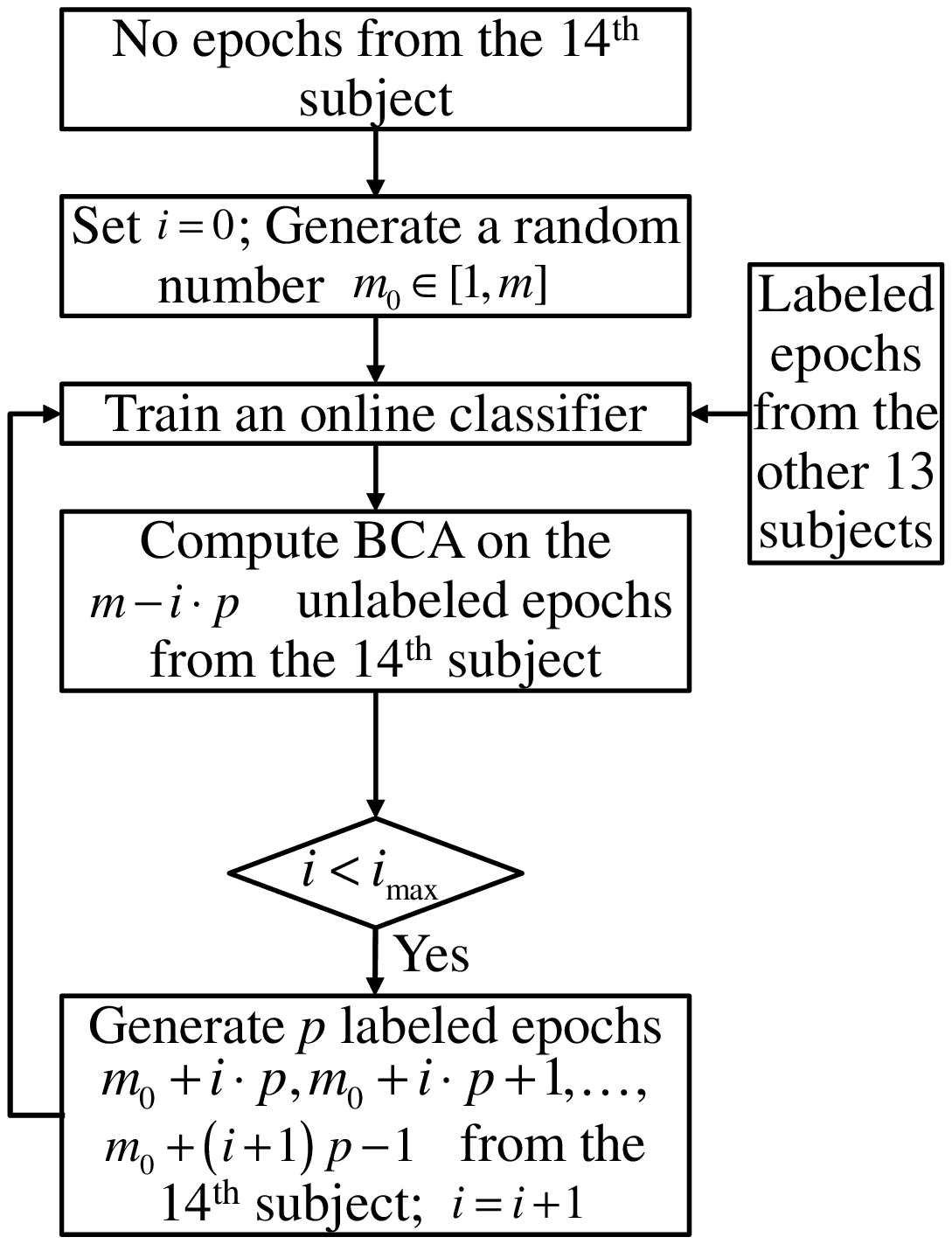}}
\caption{Flowcharts of the calibration scenarios. (a) offline; (b) online.} \label{fig:onoff}
\end{figure}

The flowchart for the simulated offline calibration scenario is shown in Fig.~\ref{fig:offlineC}. Assume the 14th subject has $m$ sequential epochs in the VEP experiment, and we want to label $p$ epochs in each iteration, starting from zero. We first generate a random number $m_0\in[1, m]$, representing the starting position in the VEP sequence. Then, in the first iteration, we use the $m$ unlabeled epochs from the 14th subject and all labeled epochs from the other 13 subjects to build different classifiers and compute their BCAs on the $m$ unlabeled epochs. In the second iteration, we obtain labels for Epochs\footnote{For offline calibration, the labeled epochs need not to be sequential: they can be randomly selected from the $m$ epochs. However, the labeled epochs are always sequential in online calibration. To facilitate the comparison between offline and online algorithms, we used sequential sampling in both offline and online calibrations. But note that there is no statistic difference between random sampling and sequential sampling in offline calibration.} $m_0$, $m_0+1$, ..., $m_0+p-1$ from the 14th subject, build different classifiers, and compute their BCAs on the remaining $m-p$ unlabeled epochs. We iterate until the maximum number of iterations is reached. When the end of the VEP sequence is reached during the iteration, we rewind to the beginning of the sequence, e.g., if $m_0=m$, then Epoch $m_0+1$ is the 1st epoch in the VEP sequence, Epoch $m_0+2$ is the 2nd, and so on.

To obtain statistically meaningful results, the above process was repeated 30 times for each subject, each time with a random starting point $m_0$. We repeated this procedure 14 times so that each subject had a chance to be the ``14th" subject.

\subsection{Algorithms} \label{sect:Aoffline}

We compared the performance of wARSDS with six other algorithms \cite{drwuSMC2015}:
\begin{enumerate}
\item \emph{BL1}, a baseline approach in which we assume we know labels of all samples from the new subject, and use 5-fold cross-validation and SVM to find the highest BCA. This represents an upper bound of the BCA we can get, by using the data from the new subject only.
\item \emph{BL2}, which is a simple iterative procedure: in each iteration we randomly select five unlabeled samples from the new subject to label, and then train an SVM classifier by 5-fold cross-validation. We iterate until the maximum number of iterations is reached.
\item \emph{TL}, which is the TL algorithm introduced in \cite{drwuSMC2014}. It simply combines the labeled samples from the new subject with samples from each existing subject and train an SVM classifier. The final classifier is a weighted average of all individual classifiers, and the weights are the corresponding cross-validation BCAs. Note that this algorithm can be used both online and offline, because it does not use any information from the unlabeled epochs.
\item  \emph{TLSDS}, which also performs SDS before the above TL algorithm.
\item \emph{ARRLS}, which was proposed in \cite{Long2014} (manifold regularization was removed), and is also the wAR algorithm introduced in Algorithm~1, by setting $w_t=w_{s,i}=w_{t,i}=1$.
\item \emph{wAR}, which excludes the SDS part in Algorithm~1.
\end{enumerate}
Weighted libSVM \cite{LIBSVM} with RBF kernel was used as the classifier in BL1, BL2, TL and TLSDS. The optimal RBF parameter was found by cross-validation. We chose $w_t=2$, $\sigma=0.1$, and $\lambda=10$, following the practice in \cite{Long2014,drwuSMC2015,drwuTNSRE2016}.

\subsection{Experimental Results} \label{sect:results}

The BCAs of the seven algorithms, averaged over the 30 runs and across the 14 subjects, are shown in Fig.~\ref{fig:offline} for the three headsets. Observe that:

\begin{figure}[htpb]\centering
\subfigure[]{\label{fig:BioSemi_offline_avgs}     \includegraphics[width=.45\linewidth,clip]{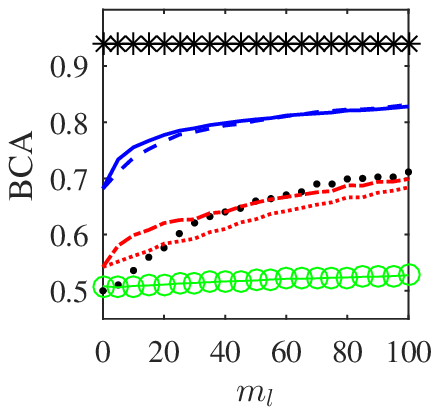}}
\subfigure[]{\label{fig:ABM_offline_avgs}     \includegraphics[width=.45\linewidth,clip]{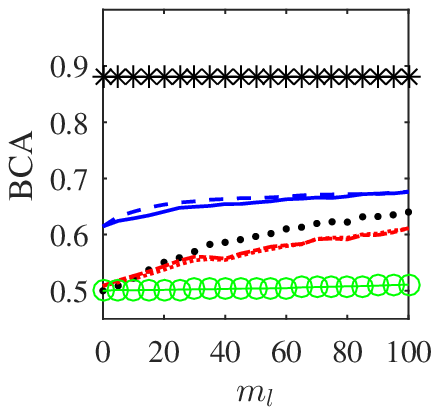}}
\subfigure[]{\label{fig:Emotiv_offline_avgs}     \includegraphics[width=.7\linewidth,clip]{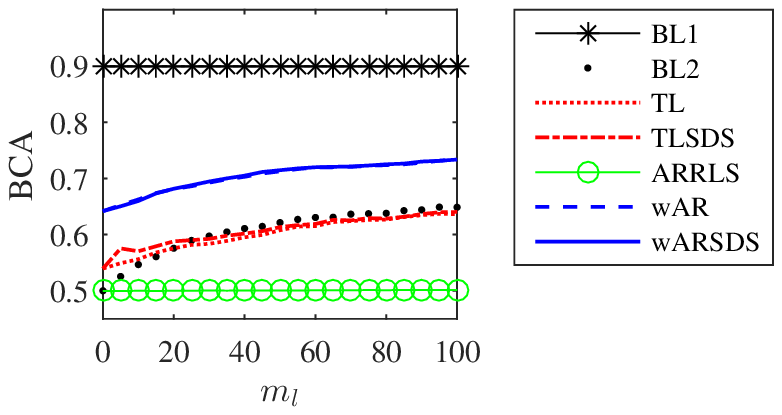}}
\caption{Average BCAs of the seven \emph{offline} algorithms across the 14 subjects, using different EEG headsets. (a) BioSemi; (b) ABM; (c) Emotiv.} \label{fig:offline}
\end{figure}

\begin{enumerate}
\item Generally the performances of all algorithms (except BL1, which is not iterative) increased as more labeled subject-specific samples were available, which is intuitive.

\item BL2 cannot build a classifier when there were no labeled subject-specific samples at all (observe that the BCA for $m_l=0$ on the BL2 curve in Fig.~\ref{fig:offline} was always $0.5$, representing random guess), but all TL/DA based algorithms can, because they can make use of information from other subjects. Moreover, without any labeled subject-specific samples, wAR and wARSDS can build a classifier with a BCA of $68.20\%$ for BioSemi, $61.45\%$ for ABM, and $64.17\%$ for Emotiv, much better than random guess.

\item Generally the performance of TL was worse than BL2, suggesting that it cannot cope well with large individual differences among the subjects\footnote{Note that this does not conflict with the observation in \cite{drwuSMC2014}, which said TL was better than BL2. This is because different datasets were used in the two studies: \cite{drwuSMC2014} downsampled the non-target class to balance the two classes before testing the performances of different algorithms, whereas class-imbalance was preserved in this paper. Moreover, \cite{drwuSMC2014} only considered the BioSemi headset, and showed that TL outperformed BL2, but the performance difference between TL and BL2 decreased as $m_l$ increases. This is also the case in Fig.~\ref{fig:BioSemi_offline_avgs}, when $m_l$ is small.}.

\item TLSDS always outperformed TL. This is because TL used a very simple way to combine the labeled samples from the new and existing subjects, and hence an existing subject whose ERPs are significantly different from the new subject's would have a negative impact on the final BCA. SDS can identify and remove (some of) such subjects, and hence benefited the BCA.

\item ARRLS demonstrated the worst BCA, because all other algorithms explicitly handled class-imbalance using weights, whereas ARRLS did not. For our dataset, the non-target class had seven times more samples than the target class, so many times ARRLS simply classified all samples as non-target, resulting in a BCA of $0.5$.

\item wAR and wARSDS significantly outperformed BL2, TL, TLSDS and ARRLS. This is because a sophisticated DA approach was used in wAR and wARSDS, which explicitly considered class imbalance, and was optimized not only for high classification accuracy, but also for small structural risk and close  feature similarity between the two domains.

\item wARSDS and wAR had very similar performance, but instead of using 13 auxiliary subjects, wARSDS only used on average 6.84 subjects for BioSemi, 6.03 subjects for ABM, and 6.85 subjects for Emotiv, corresponding to $47.38\%$, $53.62\%$ and $47.31\%$ computational cost saving, respectively.
\end{enumerate}

As in \cite{drwuSMC2015,drwuTNSRE2016}, we also performed comprehensive statistical tests to check if the performance differences among the six algorithms (BL1 was not included because it is not iterative) were statistically significant. We used the area-under-performance-curve (AUPC) \cite{drwuSMC2015,drwuRSVP2016,drwuTNSRE2016} to assess overall performance differences among these algorithms. The AUPC is the area under the curve of the BCAs obtained at each of the 30 runs, and is normalized to $[0, 1]$. A larger AUPC value indicates a better overall classification performance.

\begin{figure}[htpb]\centering
\includegraphics[width=\linewidth,clip]{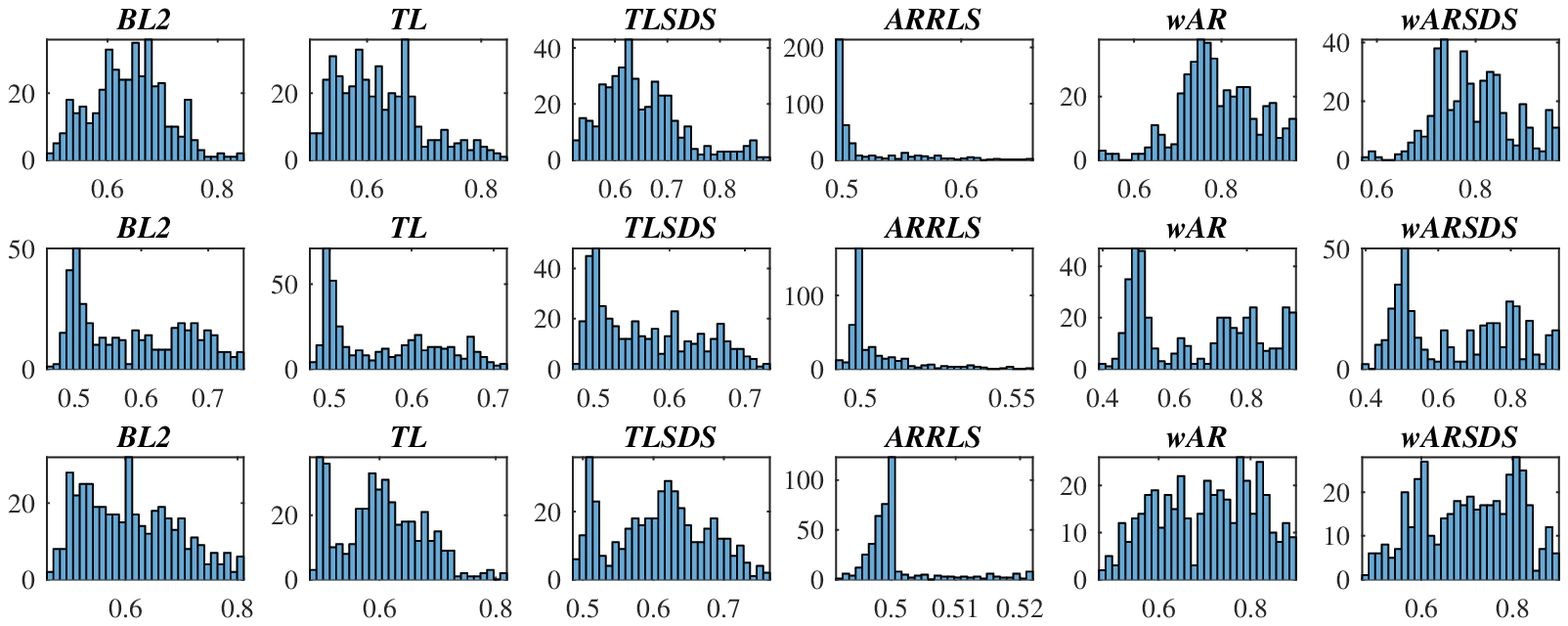}
\caption{Histograms of the AUPCs of the six algorithms on the three headsets. Top: BioSemi; middle: ABM; bottom: Emotiv.} \label{fig:hist}
\end{figure}

First, we checked the normality of our data to see if parametric ANOVA tests can be used. The histograms of the $30\times14=420$ AUPCs for each of the six algorithms on the three headsets are shown in Fig.~\ref{fig:hist}. Observe that most of them are not even close to normal. So, parametric ANOVA tests cannot be applied.

As a result, we used Friedman's test \cite{Friedman1940}, a two-way non-parametric ANOVA where column effects are tested for significant differences after adjusting for possible row effects. We treated the algorithm type (BL2, TL, TLSDS, ARRLS, wAR, wARSDS) as the column effects, with subjects as the row effects. Each combination of algorithm and subject had 30 values corresponding to 30 runs performed. Friedman's test showed statistically significant differences among the six algorithms for each headset ($df=5$, $p=0.00$).

Then, non-parametric multiple comparison tests using Dunn's procedure \cite{Dunn1961,Dunn1964} was used to determine if the difference between any pair of algorithms was statistically significant, with a $p$-value correction using the false discovery rate method \cite{Benjamini1995}. The results showed that the performances of wAR and wARSDS were statistically significantly different from BL2, TL, TLSDS and ARRLS for each headset ($p=0.0000$ for all cases, except $p=0.0031$ for ABM wAR vs BL2, and $p=0.0004$ for ABM wARSDS vs BL2). There was no statistically significant performance difference between wAR and wARSDS ($p=0.2602$ for BioSemi, $p=0.2734$ for ABM, and $p=0.4365$ for Emotiv).

In summary, we have demonstrated that given the same number of labeled subject-specific training samples, wAR and wARSDS can significantly improve the offline calibration performance. In other words, given a desired classification accuracy, wAR and wARSDS can significantly reduce the number of labeled subject-specific training samples. For example, in Fig.~\ref{fig:BioSemi_offline_avgs}, the average BCA of BL2 is $71.14\%$, given 100 labeled subject-specific training samples. However, to achieve that BCA, on average wAR and wARSDS only need 5 samples, corresponding to $95\%$ saving of the labeling effort. Moreover, Fig.~\ref{fig:BioSemi_offline_avgs} also shows that, without using any labeled subject-specific samples, wAR and wARSDS can achieve similar performance as BL2 which uses 65 samples. Similar observations can also be made for the ABM and Emotiv headsets.

\subsection{Parameter Sensitivity Analysis}

In this subsection we study the sensitivity of wAR and wARSDS to parameters $\sigma$ and $\lambda_P$ ($\lambda_Q$). To save space, we only show the BCA results for the BioSemi headset. Similar results were obtained from the other two headsets.

\begin{figure}[htpb]\centering
\subfigure[]{\label{fig:sigma}\includegraphics[width=.9\linewidth,clip]{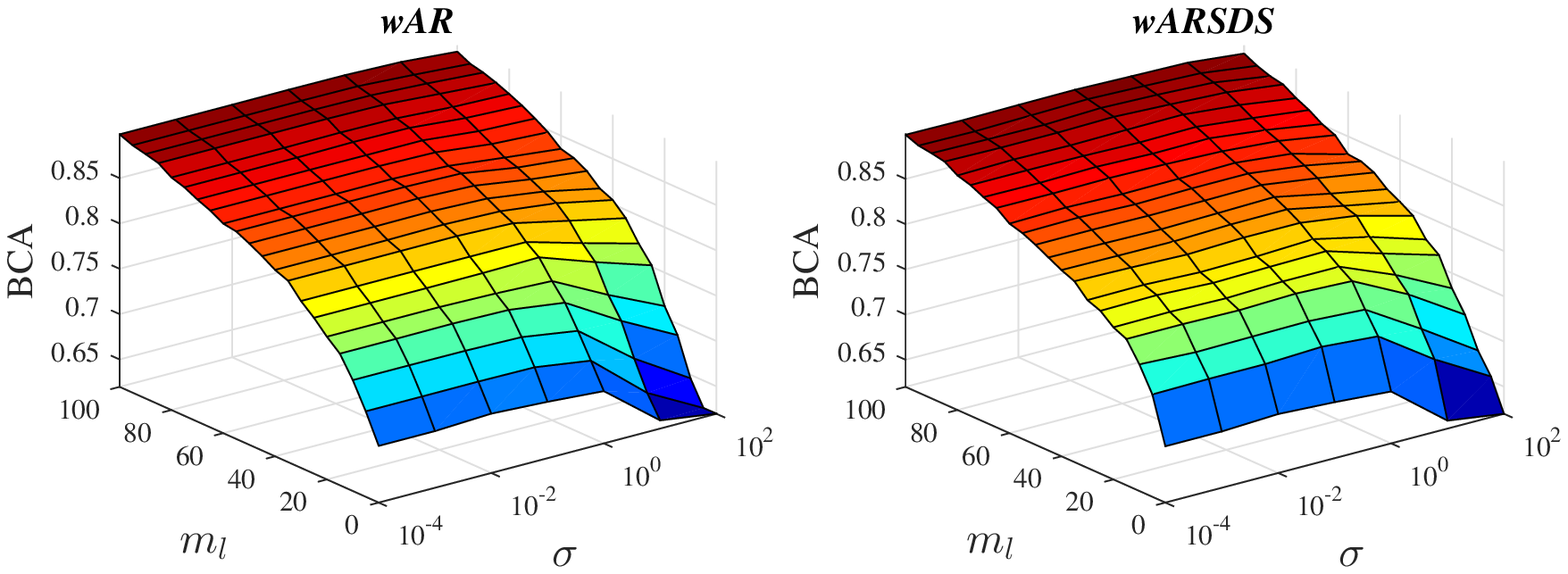}}
\subfigure[]{\label{fig:lambda}\includegraphics[width=.9\linewidth,clip]{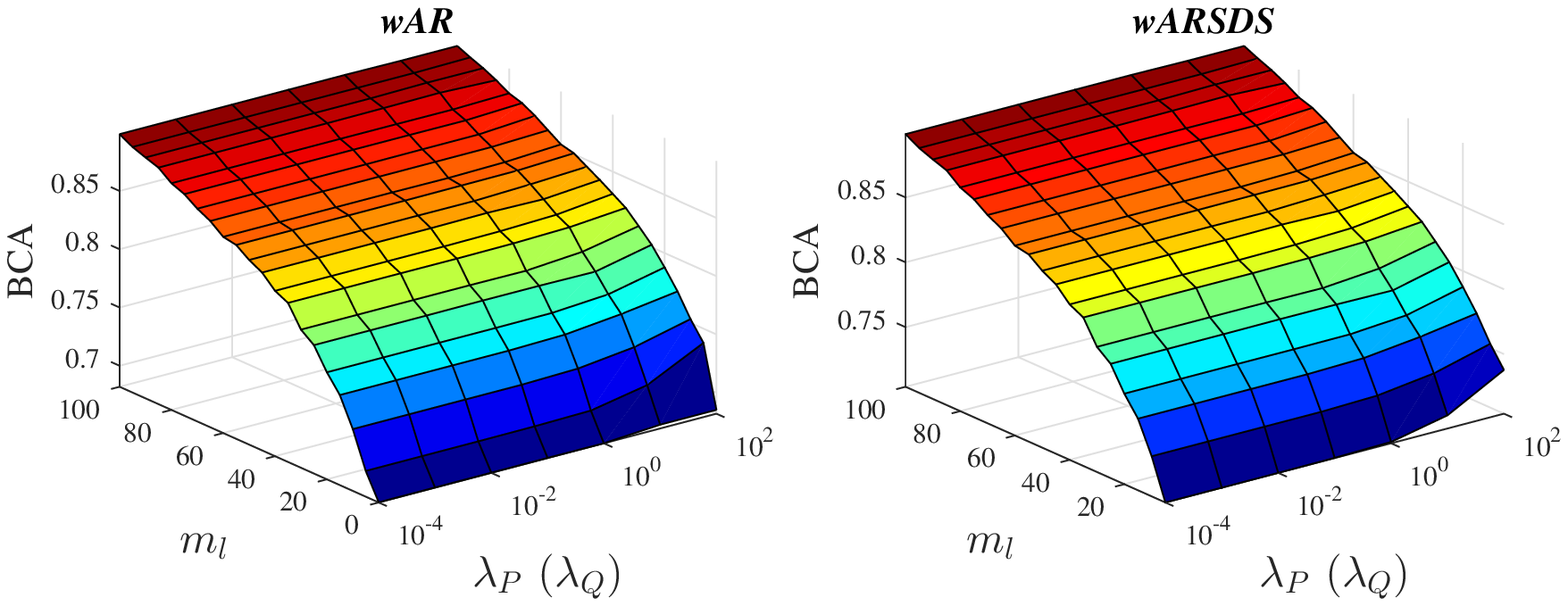}}
\caption{Average BCAs of wAR and wARSDS for different parameters for the BioSemi headset. (a) $\sigma$; and, (b) $\lambda_P$ and $\lambda_Q$.} \label{fig:parameters}
\end{figure}

The average BCAs of wAR and wARSDS for different $\sigma$ ($\lambda_P$ and  $\lambda_Q$ were fixed at 10) are shown in Fig.~\ref{fig:sigma}, and for different $\lambda_P$ and\footnote{We always assigned $\lambda_P$ and $\lambda_Q$ identical value because they are conceptually close.} $\lambda_Q$ ($\sigma$ was fixed at 0.1) are shown in Fig.~\ref{fig:lambda}.  Observe from Fig.~\ref{fig:sigma} that both wAR and wARSDS achieved good BCAs for $\sigma\in[0.0001, 1]$, and from Fig.~\ref{fig:lambda} that both wAR and wARSDS achieved good BCAs for $\lambda_P\in[10, 100]$ and $\lambda_Q\in[10,100]$. Moreover, $\sigma$, $\lambda_P$ and $\lambda_Q$ have more impact to the BCA when $m_l$ is small. As $m_l$ increases, the impact diminishes. This is intuitive, as the need for transfer diminishes as the amount of labeled target domain data increases.

\section{Performance Evaluation of Online Calibration Algorithms} \label{sect:online}

This selection compares the performance of OwARSDS with several other online calibration algorithms.

\subsection{Online Calibration Scenario}

Although we knew the labels of all EEG epochs for all 14 subjects in the experiment, we simulated a realistic online calibration scenario: we had labeled EEG epochs from 13 subjects, but initially no epoch from the 14th subject; we generated labeled epochs from the 14th subject iteratively and sequentially on-the-fly, which were used to train a classifier to label the remaining epochs from that subject.

The flowchart for the simulated online calibration scenario is shown in Fig.~\ref{fig:onlineC}. Compared with the offline calibration scenario in Fig.~\ref{fig:offlineC}, the main difference is that offline calibration has access to all $m$ unlabeled samples from the 14th subject, but online calibration does not.

More specifically, assume the 14th subject has $m$ sequential epochs in the VEP experiment, and we want to label $p$ epochs in each iteration, starting from zero. We first generate a random number $m_0\in[1, m]$, representing the starting position in the VEP sequence. Then, in the first iteration, we use all labeled epochs from the other 13 subjects to build different classifiers, and compute their BCAs on the $m$ unlabeled epochs. In the second iteration, we generated labeled Epochs $m_0$, $m_0+1$, ..., $m_0+p-1$ from the 14th subject, build different classifiers, and compute their BCAs on the remaining $m-p$ unlabeled epochs. We iterate until the maximum number of iterations is reached. To obtain statistically meaningful results, the above process was repeated 30 times for each subject, each time with a random starting point $m_0$. The whole process was repeated 14 times so that each subject had a chance to be the ``14th" subject.

\subsection{Online Calibration Algorithms}

We compared the performances of OwARSDS with five other algorithms:
\begin{enumerate}
\item \emph{BL1} in Section~\ref{sect:Aoffline}.
\item \emph{BL2} in Section~\ref{sect:Aoffline}, using different PCA features.
\item \emph{TL} in Section~\ref{sect:Aoffline}, using different PCA features.
\item  \emph{TLSDS}, which is the above TL algorithm with SDS.
\item \emph{OwAR}, which uses all existing subjects, instead of performing SDS.
\end{enumerate}
Again, weighted libSVM \cite{LIBSVM} with RBF kernel was used as the classifier in BL1, BL2, TL and TLSDS. We chose $w_t=2$, $\sigma=0.1$, and $\lambda=10$.

Note that the online algorithms still used PCA features, but they were computed differently from those in offline calibration. In offline calibration we had access to the $m_l$ labeled samples plus the $m_u$ unlabeled samples, so the PCA bases can be pre-computed from all $m_l+m_u$ samples and kept fixed in each iteration. However, in online calibration, we only had access to the $m_l$ labeled samples, so the PCA bases were computed from the $m_l$ samples only, and we updated them in each iteration as $m_l$ changed.

\subsection{Experimental Results} \label{sect:OnlineResults}

\begin{figure}[htpb]\centering
\subfigure[]{\label{fig:BioSemi_online_avgs}     \includegraphics[width=.45\linewidth,clip]{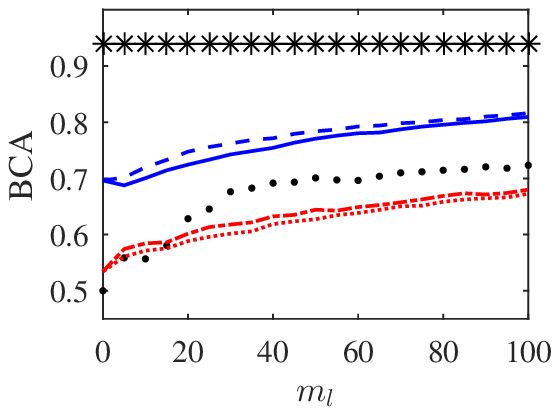}}
\subfigure[]{\label{fig:ABM_online_avgs}     \includegraphics[width=.45\linewidth,clip]{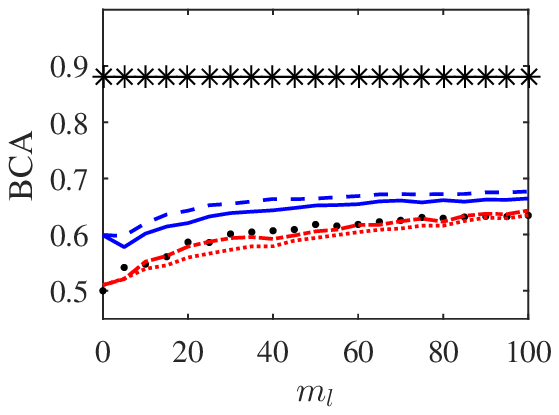}}
\subfigure[]{\label{fig:Emotiv_online_avgs}     \includegraphics[width=.7\linewidth,clip]{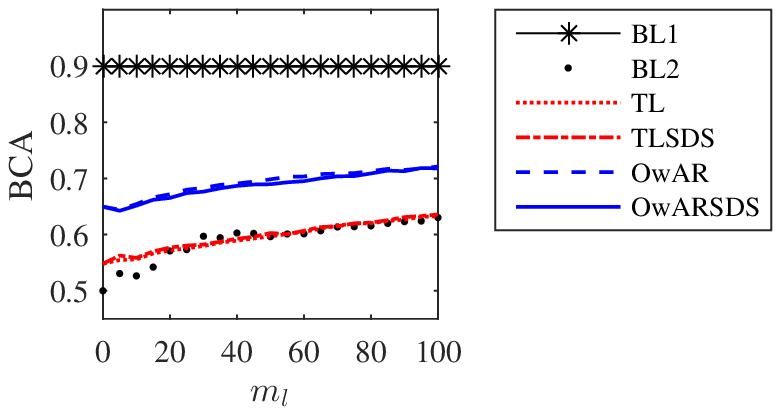}}
\caption{Average BCAs of the six \emph{online} algorithms across the 14 subjects, using different EEG headsets. (a) BioSemi; (b) ABM; (c) Emotiv.} \label{fig:online}
\end{figure}

The BCAs of the six algorithms, averaged over the 30 runs and across the 14 subjects, are shown in Fig.~\ref{fig:online} for different EEG headsets. The observations made in Section~\ref{sect:results} for offline calibration still hold here, except that ARRLS was not included in online calibration. Particularly, both OwAR and OwARSDS achieved much better performance than BL2, TL and TLSDS. However, instead of using 13 auxiliary subjects in OwAR, OwARSDS only used on average 6.51 subjects for BioSemi, 6.01 subjects for ABM, and 7.09 subjects for Emotiv, corresponding to $49.92\%$, $53.77\%$ and $45.46\%$ computational cost saving, respectively.

Friedman's test showed statistically significant performance differences among the five algorithms (excluding BL1, which is not iterative) for each headset ($df=4$, $p=0.00$). Dunn's procedure showed that the BCAs of OwAR and OwARSDS are statistically significantly different from BL2, TL, and TLSDS for each headset ($p=0.0002$ for ABM OwARSDS vs BL2, $p=0.0001$ for ABM OwARSDS vs TLSDS, and $p=0.0000$ in all other cases). There was no statistically significant performance difference between OwAR and OwARSDS ($p=0.0682$ for BioSemi, $p=0.1929$ for ABM, and $p=0.3554$ for Emotiv).

In summary, we have demonstrated that given the same number of labeled subject-specific training samples, OwAR and OwARSDS can significantly improve the online calibration performance. In other words, given a desired classification accuracy, OwAR and OwARSDS can significantly reduce the number of labeled subject-specific samples. For example, in Fig.~\ref{fig:BioSemi_online_avgs}, the average BCA of BL2 was $72.34\%$, given 100 labeled subject-specific training samples. However, to achieve that BCA, on average OwAR only needed 15 samples, and OwARSDS only needed 20 samples, corresponding to 85\% and 80\% saving of labeling effort, respectively. Moreover, Fig.~\ref{fig:BioSemi_online_avgs} also shows that, without using any labeled subject-specific samples, OwAR and OwARSDS can achieve similar BCA as BL2 which used 60 labeled subject-specific samples. Similar observations can also be made for the ABM and Emotiv headsets.

\section{Comparison of Offline and Online Algorithms} \label{sect:onoff}

This section compares the performances of wARSDS and OwARSDS (wAR and OwAR). Intuitively, we expect the performances of the offline calibration algorithms to be better than their online counterparts, because: 1) offline calibration uses all $m_l+m_u$ EEG epochs to compute the PCA bases, whereas online calibration only uses $m_l$ epochs, so the PCA bases in offline calibration should be more representative; and, 2) offline calibration also uses the $m_u$ unlabeled epochs in the optimization, whereas online calibration does not, so offline calibration makes use of more information. In other words, offline calibration makes use of semi-supervised learning whereas online calibration does not.

The average performances of wAR, wARSDS, OwAR and OwARSDS across the 14 subjects are shown in Fig.~\ref{fig:onoff}. Observe that the results were consistent with our expectation: for all three headsets, the offline algorithms (wAR and wARSDS) achieved better BCAs than their online counterparts (OwAR and OwARSDS). Additionally, Fig.~\ref{fig:onoff} shows that the algorithms had best performance using the BioSemi headset, and worst performance using the ABM headset. This is not surprising, as BioSemi used the most number of channels, and it was wired, which means better signal quality. The ABM headset had the least number of channels, and was wireless. Moreover, epochs with incorrect button presses were filtered out for BioSemi and Emotiv headsets, but not for most subjects for the ABM headset. So, the epochs in ABM were noisier.

We also performed statistical tests to check if the performance differences among the four algorithms were statistically significant. Friedman's test showed statistically significant differences among the four algorithms for BioSemi ($df=3$, $p=0.00$) and Emotiv ($df=3$, $p=0.04$), but not ABM ($df=3$, $p=0.38$). Dunn's procedure showed that for BioSemi the BCAs of wAR and OwAR were statistically significantly different ($p=0.0043$), so were BCAs of wARSDS and OwARSDS ($p=0.0000$). For ABM and Emotiv the performance differences between online and offline algorithms were not statistically significant ($p=0.5043$ for ABM wAR vs OwAR, $p=0.1959$ for ABM wARSDS vs OwARSDS, $p=0.0838$ for Emotiv wAR vs OwAR, and $p=0.0514$ for Emotiv wARSDS vs OwARSDS).

In conclusion, we have shown that generally the offline wAR and wARSDS algorithms, which include a semi-supervised learning component, can achieve better calibration performance than the corresponding online OwAR and OwARSDS algorithms, i.e., semi-supervised learning is effective.

\begin{figure}[htpb]\centering
\subfigure[]{\label{fig:BioSemi_onoff_avgs}     \includegraphics[width=.45\linewidth,clip]{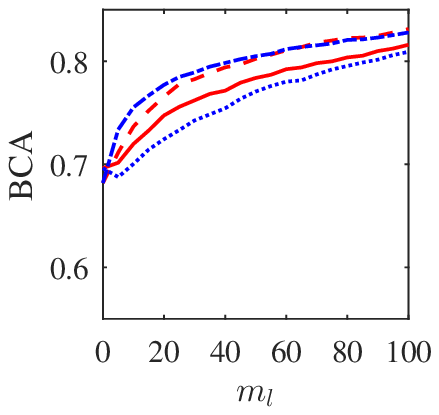}}
\subfigure[]{\label{fig:ABM_onoff_avgs}     \includegraphics[width=.45\linewidth,clip]{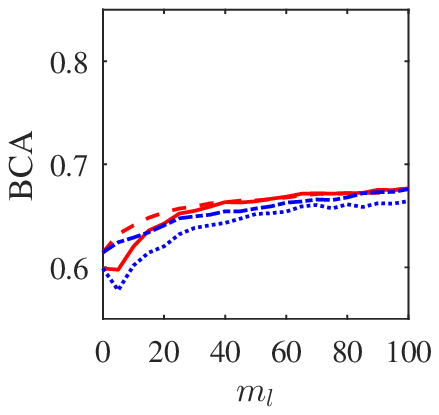}}
\subfigure[]{\label{fig:Emotiv_onoff_avgs}     \includegraphics[width=.7\linewidth,clip]{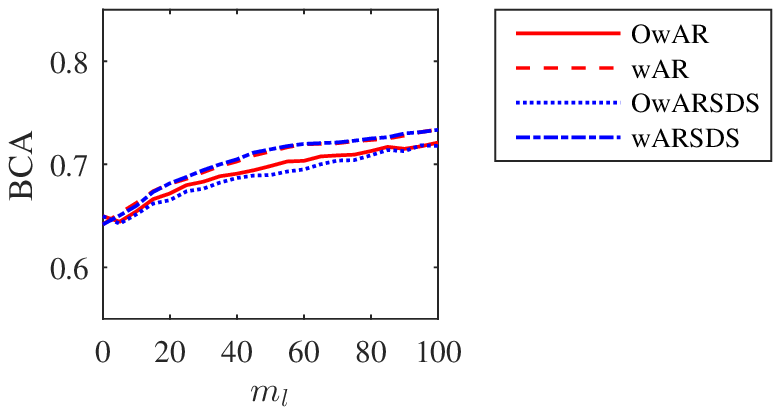}}
\caption{Average BCAs of wAR, wARSDS, OwAR and OwARSDS across the 14 subjects, using different EEG headsets. (a) BioSemi; (b) ABM; (c) Emotiv.} \label{fig:onoff}
\end{figure}

\section{Conclusions} \label{sect:conclusions}

Single-trial classification of ERPs in EEG signals is used in many BCI applications. However, because different subjects have different neural responses to even the same stimulus, it is very difficult to build a generic ERP classifier whose parameters fit all subjects. So, the classifier needs to be calibrated for each individual subject, using some labeled subject-specific data. Reducing this calibration effort, i.e., minimizing the number of labeled subject-specific data required in the calibration, would greatly increase the utility of a BCI system. This paper introduced both online and offline wAR algorithms for this purpose. We have demonstrated using a VEP oddball task and three different EEG headsets that both algorithms can cope well with the class-imbalance problem, which is intrinsic in many real-world BCI applications, and they also significantly outperformed several other algorithms. We also compared the performances of the online and offline wAR algorithms in identical experimental settings and showed that the offline wAR algorithm, which includes an extra semi-supervised component than the online wAR algorithm, can achieve better calibration performance, i.e., semi-supervised learning is effective in BCI calibration. Moreover, we further proposed a source domain selection approach, which can reduce the computational cost of both online and offline wAR algorithms by about $50\%$.

We expect our algorithms to find broad applications in various BCI calibration scenarios, and beyond. The most intuitive BCI calibration scenario, as described in this paper, is to reduce the subject-specific calibration data requirement by making use of relevant data from other subjects. Another scenario is to make use of the same subject's data from previous usages to facilitate a new calibration. For example, the subject may need to work on the same BCI task at different locations using different EEG headsets (office, home, etc.), or may upgrade a BCI game with a new EEG headset. In such applications, wAR can be used to make use of the data obtained from a previous EEG headset to facilitate the calibration for the new headset, as introduced in \cite{drwuTNSRE2016}. Of course, the above two scenarios can also be combined: auxiliary data from other subjects and from the subject himself/herself can be integrated to expedite the calibration. Furthermore, because of human-machine mutual adaptation and non-stationarity, a well-calibrated BCI system may degrade gradually. The proposed wAR algorithms can be used to re-calibrate it from time to time. Additionally, EEGs, together with many other body signals (facial expressions, speech, gesture, galvanic skin response, etc.), are also frequently used in affective computing \cite{Picard1997}, ``\emph{computing that relates to, arises from, or deliberately influences emotion or other affective phenomena.}" The wAR algorithms can also be used to handle individual differences and non-stationarity in such applications.

Finally, we need to point out that the current wAR algorithms still have some limitations, which will be improved in our future research. First, we will develop incremental updating rules to reduce their computational cost. Second, we will develop criteria to determine when a negative transfer may occur, and hence use subject-only calibration data in such cases. Third, although wAR can map the features to a new kernel space to make them more consistent across the source and target domains, it still relies on good initial features. Simple PCA features were used in this paper. In our future research we will consider more sophisticated and robust features, e.g., the information geometry \cite{Barachant2012}.

\section*{Acknowledgement}

The author would like to thank Scott Kerick, Jean Vettel, Anthony Ries, and David W. Hairston at the U.S. Army Research Laboratory (ARL) for designing the experiment and collecting the data, and Brent J. Lance and Vernon J. Lawhern from the ARL for helpful discussions.

\textit{Research was sponsored by the U.S. Army Research Laboratory and was accomplished under Cooperative Agreement Numbers W911NF-10-2-0022 and W911NF-10-D-0002/TO 0023. The views and the conclusions contained in this document are those of the authors and should not be interpreted as representing the official policies, either expressed or implied, of the U.S. Army Research Laboratory or the U.S Government.}



\end{document}